\documentclass[10pt,twocolumn,letterpaper]{article}

\usepackage{iccv}
\usepackage{times}
\usepackage{epsfig}
\usepackage{graphicx}
\usepackage{amsmath}
\usepackage{amssymb}
\usepackage{supertabular,enumitem}
\usepackage{sidecap}

\usepackage{xcolor}
\usepackage{colortbl}
\usepackage{rotating}
\usepackage{booktabs}
\usepackage{multirow}
\usepackage{multicol}

\usepackage[nodisplayskipstretch]{setspace}
\usepackage[ruled,vlined]{algorithm2e}

\SetCommentSty{mycommfont}

\SetKwInput{KwInput}{Input}                
\SetKwInput{KwOutput}{Output}              

\newcommand{\squeezeup}{\vspace{-2.5mm}}
\newcommand{\comment}[1]{}

\usepackage[pagebackref=true,breaklinks=true,letterpaper=true,colorlinks,bookmarks=false]{hyperref}

\iccvfinalcopy 

\def\iccvPaperID{5976} 

\begin{document}

\title{Physical world assistive signals for deep neural network classifiers - neither defence nor attack}

\author{
Camilo Pestana, Wei Liu, David Glance, Robyn Owens, Ajmal Mian \\
Department of Computer Science\\
The University of Western Australia\\
35 Stirling Hwy, Crawley WA 6009, Australia\\
Email: camilo.pestanacardeno@research.uwa.edu.au, \\ \{wei.liu, david.glance, robyn.owens, ajmal.mian\}@uwa.edu.au
}

\maketitle
\ificcvfinal\thispagestyle{empty}\fi

\begin{abstract}
   Deep Neural Networks lead the state of the art of computer vision tasks. Despite this, Neural Networks are brittle in that small changes in the input can drastically affect their prediction outcome and confidence. Consequently and naturally, research in this area mainly focus on adversarial attacks and defenses. In this paper, we take an alternative stance and introduce the concept of \textbf{Assistive Signals}, which are optimized to improve a model's confidence score regardless if it's under attack or not. We analyse some interesting properties of these assistive perturbations and extend the idea to optimize assistive signals in the 3D space
   for real-life scenarios simulating different lighting conditions and viewing angles. Experimental evaluations show that the assistive signals generated by our optimization method increase the accuracy and confidence of deep models more than those generated by conventional methods that work in the 2D space.
   Our `Assistive Signals' illustrate the intrinsic bias of ML models towards certain patterns in real-life objects. We discuss how we can exploit these insights to re-think, or avoid, some patterns that might contribute to, or degrade, the detectability of objects in the real-world. 
\end{abstract}

\section{Introduction}

\begin{figure}[h!]
   \centering
   \includegraphics[width=\linewidth]{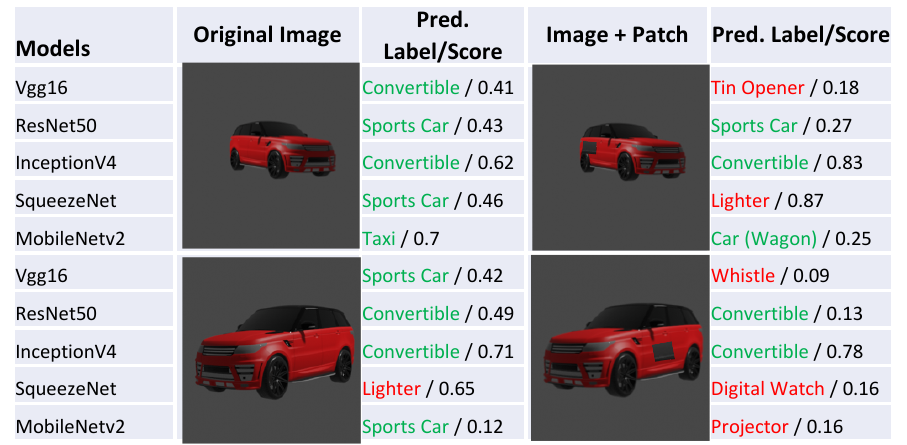} 
   \caption{
   We arbitrarily added a non-adversarial black patch to render images of a car for different ImageNet-based models. Given that different models' predictions differed with respect to the specific label, we consider as a valid label (green color) all the predicted classes that are related to a car and wrong labels (red color) those that are significantly different from a car or transport vehicle.
   }
\label{fig:occlusion}
\vspace{3mm}
\end{figure}

The study of adversarial examples in computer vision is often motivated by using human indistinguishable small changes in an input image to cause model prediction errors~\cite{szegedy2014intriguing,gu2015deep,nguyen2015deep,elsayed2018adversarial}. In this threat scenario, the adversary is allowed to perturb all pixels in an image such that the $\ell_p$-norm of the perturbation is constrained to be within a prescribed bound. However, many of these attacks are shown to be rather inefficient in real-life scenarios ~\cite{chiang2020certified,liu2019dpatch}. Adversarial patch attacks \cite{adversarialpatch2017,liu2019dpatch,li2019exploring}, where patches of certain patterns are added to input images, are among the most practical threats against real-world computer vision systems.

Consequently, research in adversarial machine learning (ML) has been focused around the idea of creating more efficient attacks, better defenses and exploring the concepts of what makes ML models more robust against those attacks.
In this paper, we look at the problem on how a physical object or its image can be guaranteed to be correctly recognized. Perturbations on the input images or alterations in the object are used to improve the prediction confidence of the correct class through full Assistive Perturbations or Assistive Patches. 

The term ‘Assistive Signals’ is not defined in the existing literature as it is a novel research direction. Hence, we coined this term referring to any embedded signal in the input that enhances the confidence score of the prediction made by a classifier to recover with ‘good’ confidence the correct class label under natural or intentional adversarial changes in the environment. For example, if a 3D object (e.g, car) is attacked by an adversary modifying scene properties such as object pose or lighting so that it is misclassified by a model, then ‘Assistive Signals’ embedded in the 3D object should ideally ‘assist’ the model to correctly classify the object despite the adversarial conditions. ‘Assisting’ a ML classifier is the main purpose of the \textit{Assistive signals}, and such signals can be expressed in many forms including but not limited to \textit{Assistive Textures} and \textit{Assistive Patches}. Therefore, the term `Assistive Signals' confers enough meaning to give an intuition about its purpose and is also sufficiently generic to include all future work in regards to the many types of assistive signals that could be created in this emerging area of research. It is important to notice that that concept of `Assistive Signals' might seem similar to adversarial attacks in that both approaches can perturb or alter the input. It can also appear similar to defenses in that both could improve the classifier's confidence scores under attack. However, despite the similarities between these concepts, there are some key differences:

\begin{figure}[t!]
   \centering
   \includegraphics[width=\linewidth]{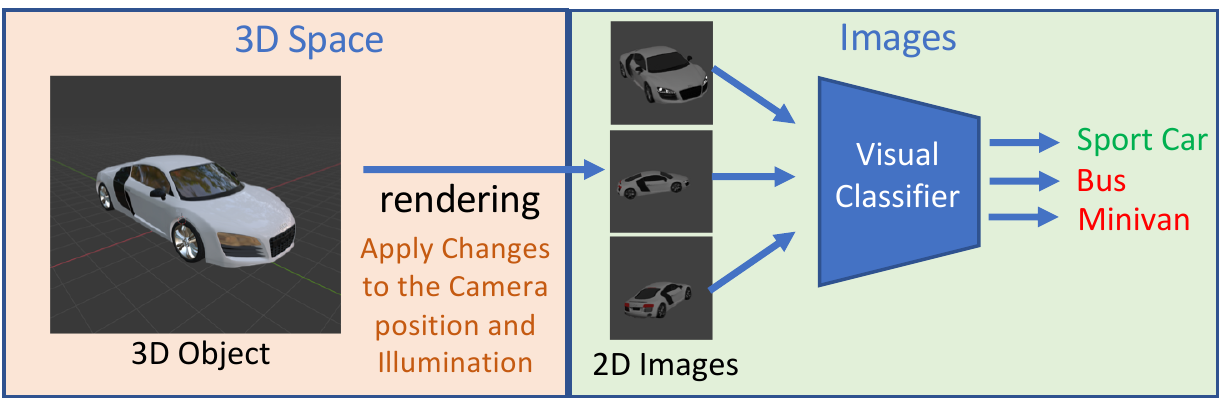}
   \caption{
Most adversarial attacks focus on modifying pixels in 2D images to fool CNN models. 
According to Zeng et al.~\cite{Zeng_2019_CVPR}, 2D images are projections of the underlying 3D scene suggesting that adversarial attacks can go beyond the image space. In this example, we apply some changes in the lighting and camera position which are sufficient to make the rendered images of the 3D object to be missclassified by a visual classifier (Vgg16 here).}
\label{fig:3dspaceattacks}
\vspace{3mm}
\end{figure}

First, Assistive Signals are meant to improve the general ability of classifiers to recognize the target object predicting the correct class (in some cases with higher confidence) whether the classifier is under attack or not. Most defenses focus in the recovery of the accuracy only when the classifier is under attack while degrading their performance in `normal' conditions.

Second, current literature in adversarial ML focus on the software/algorithmic side, where attack or defence techniques are developed for. Instead, the assistive signals are operating in the physical space, i.e. how we alter the appearance of an object to protect it from being misclassified (e.g. a car), or camouflage an object to protect it from being classified (e.g. a submarine or tank).

Third, instead of relying on the predictive capabilities of a classifier, Assistive Signals make the physical object itself more robust against adversarial conditions and easier to classify under `normal' circumstances. We show that our `Assistive Signals' transfer well to other classifiers. As an example of future applications, if a self-driving manufacturer company improves the design of their cars to be more `detectable' using assistive signals, those same modifications will help third party companies to classify such cars with higher confidence. Hence, instead of autonomous vehicle manufactures working on improving their own computer vision models, they would be making their cars more recognizable in general, agnostic to any computer vision system. Such a simple enhancement would have a major impact on the road safety of self-driving cars in the near future. This is just an example of how practical this new research direction could become.

\begin{figure}[t!]
   \centering
   \scriptsize
   \includegraphics[width=\linewidth]{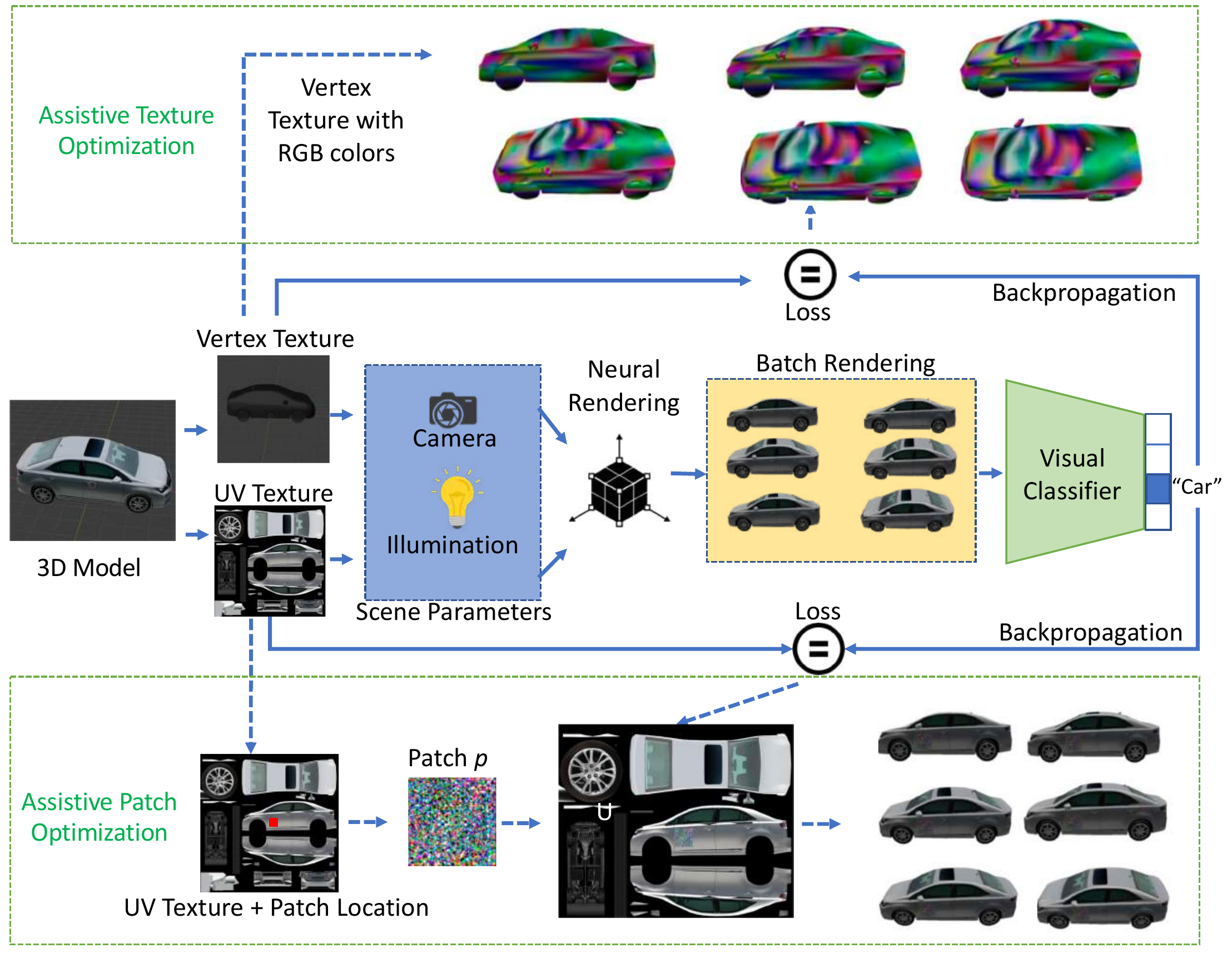} 
   
   \caption{
   Using a 3D model, our optimization method iteratively generates a full Assistive Texture or an Assistive Patch $\rho$ for a visual classifier (DenseNet121 here). In each step, the perturbation is simultaneously optimized for different viewing angles and lighting conditions making our assistive signals invariant.}
\label{fig:differentiable_assistive}
\vspace{3mm}
\end{figure}

Lastly, while the focus of this paper is in the physical alteration of man-made 3D objects, the more general concept of Assistive Signals 
has a broader scope. For example, future research in this area could explore the impact of Assistive Signals in different applications and types of data such as video and sound, i.e. how to improve human action recognition from video using assistive signals.
\\
\indent Further, our biggest motivation for the creation of `Assistive Signals' embedded into objects is the fact that fooling a model in the 3D scene is easier than making it robust, as demonstrated by Zeng et al.~\cite{Zeng_2019_CVPR}. Hence, Assistive Signals are a more challenging task than creating adversarial signals.\\
\\
Figure~\ref{fig:occlusion} illustrates how even a simple non-adversarial black patch (i.e.~occlusion) on the 3D object can change the confidence scores of different ImageNet-based~\cite{krizhevsky2012ImageNet} models. Moreover, Figure~\ref{fig:3dspaceattacks} shows how changing the camera views and lighting conditions of the scene are enough to fool an image classifier. Therefore, the creation of crafted adversarial attacks is not even necessary to fool a model when simple changes of the camera view, illumination and occlusions can pose a significant threat to the model. These challenges can be addressed using Assistive Signals by improving the object's design using the insights from \textit{Robust Patterns}, or modifying the appearance of existing physical objects with Assistive Textures and Patches.
\\
\indent Figure \ref{fig:differentiable_assistive} illustrates the intuition of how assistive patches and textures are generated with the aim of preserving the classification outcome using batch neural rendering. This is in contrast to the conventional adversarial examples which are perturbations meant to deceive a model from making the correct prediction. In the context of this research, we term such attack oriented perturbations as \emph{Deceptive Signals}. The main contributions of this paper are thus fourth-fold:

\begin{enumerate}[leftmargin=*,label=$\bullet$,noitemsep,partopsep=0pt,topsep=0pt,parsep=0pt]
    \item We propose the first `adversarial' learning technique that optimizes perturbations in the input for the purpose of \emph{improving} the confidence score of an image classification model instead of degrading its performance\footnote{Code Available at: https://github.com/elcronos/assistive-signals}. 
\item We propose an algorithm that can generate \emph{Assistive} and \emph{Deceptive} textures and patches that are invariant to 3D physical properties such as the camera orientation and lighting conditions etc. 
\item We propose two types of Assistive Signals for improving the detectability of 3D objects. First, Assistive Patches can be thought as a retrofit solution that can be added to existing physical objects. On the other hand, Assistive Textures can be embedded at the moment of creation of such objects.

\item We introduce the concept of \textit{Robust Patterns} that can be extracted from our Assistive Signals. Such signals provide critical insights (e.g., to manufacturers of autonomous cars) on what robust patterns such as color, appearance, and semantic structure could make 3D objects such as Cars, Stop Signs, Traffic Lights more readily recognizable to a deep learning model.
\end{enumerate}

\section{Problem Definition}
\noindent Let  $x$ $\in$ $\mathbb{R}^{H \times W \times 3}$ denote an image with height $H$, width $W$ and three color channels, usually (R)ed, (G)reen and (B)lue. An image $\rho$ is a \textbf{perturbed image} if $d(x, x+\rho)$ $< \epsilon$, where $d$ is a distance metric and $\epsilon$ is a small number. 
Let $y = \{p_1, …, p_m\} \in R^n$ be the probabilities of the predicted labels {1,..., $m$}. If $\mathcal{C}$ is an image classifier, then $\mathcal{C}:x \rightarrow \{{k(x), p_k(x)}\}$, where $k(x)$ is the label of $x$ with the highest confidence $p_k(x)$. An \textbf{assistive image} is defined as $x_{asst} = x + \rho_{asst}$, such that if $\mathcal{C}(x) = \{{k(x), p_k(x)}\}$ and $\mathcal{C}(x_{asst}) = \{{k(x_{asst}), p_k(x_{asst})}\}$, then $k(x) = k(x_{asst})$ and $p_k(x_{asst}) > p_k(x)$. In other words, the classifier returns the same correct label with a higher confidence. 

\noindent In the context of deceptive signals, adversarial attacks \cite{akhtar2018threat} are defined as $x_{adv} =  x+ \rho_{adv}$, where the adversarial perturbation $\rho_{adv}$ is chosen to ensure that if $\mathcal{C}: x \rightarrow \{{k(x), p_k(x)}\}$ then $k(x) \neq k(x_{adv})$. After an adversarial attack, the classifier returns the incorrect label. The calculation $x_{asst} = x +\rho_{asst}$ can be constrained such that the perturbation in $x_{asst}$ is imperceptible to the human eye, that is $d(x, x_{asst}) < \epsilon$ for a small value of $\epsilon$. This is usually the case when the perturbation is adversarial rather than assistive. However, given that an assistive perturbation is specifically created to facilitate the prediction of a class of the image, in most realistic scenarios assistive perturbations do not necessarily have to be subtle or constrained by $\epsilon$.

\section{Related Work}
We briefly review attack techniques to provide an overview for image perturbation algorithms. Additionally, we discuss other relevant topics to our work such as neural rendering, which is the underlaying technique used for the optimization of the assistive signals in the 3D space.

\subsection{Adversarial Perturbation Algorithms}
Gradient-based attacks are more powerful than non-gradient-based attacks and usually less computationally expensive \cite{akhtar2018threat}. We use gradient-based attacks to test and compare our hardened datasets using assistive signals.

\indent Fast Gradient Signed Method (FGSM)~\cite{goodfellow2014explaining}: is calculated as $x_{adv} = x + \varepsilon * \textrm{sign}(\nabla x J(\theta, x, y))$,  where  $x_{adv}$ is the adversarial image and $x$ is the original image.  In the cost function $J(\theta, x, y)$, $\theta$ represents the network parameters and $y$ the ground truth label. Moreover, $\epsilon$ is used to scale the noise and is usually a small number (e.g. $\epsilon$=0.01).\\
\indent Projected Gradient Descent (PGD) attack: is considered one of the strongest attacks and is used as a benchmark to measure the robustness of many defenses in the literature~\cite{madry2017towards}. It works similarly to FGSM, however, in this iterative version a small perturbation step $\alpha$ is applied and this increases the success rate of the  generated adversarial images. 
\\
\indent Adversarial Patch: Brown et al. \cite{brown2018adversarial} presented an adversarial attack that that works well across different scene and under a wide variety of transformations. They demonstrated the feasibility of this attack in the real world by printing these patches, adding them to any scene and then photographing the patched scene for new images to fool a classifier. Thus, Liu et al. \cite{liu2019dpatch} proposed DPatch to extend the adversarial patches to object detection. However, in this paper, we focus on the image classification problem.
\\
\indent Expectation-over-transformation (EoT):
Athalye et al.~\cite{athalye2018synthesizing} proposed training 2D images and 3D objects with a transformation distribution using different camera distances, lighting
conditions, translation and rotation of the object to create more resilient adversarial examples to those transformations. Currently, many adversarial attacks aim to work better in the physical world by implementing EoT. For example, CAMOU~\cite{zhang2018camou} attack uses this approach to improve the fooling rate.
\\
\indent AdvCam \cite{duan2020adversarial}: Duan et al. proposed an interesting approach aiming to hide large perturbations into customized styles for physical-world objects. In their \textit{adversarial camouflage loss}, they combined an adversarial loss with a camouflage loss (style loss, content loss and smoothness loss). In their approach they use 2D images (non-rendered images) and focus on crafting stealthy attacks.
\\
\indent Strike (with) a pose: Alcorn et al.~\cite{alcorn2019strike} presented a framework for discovering adversarial positions on rendered 3D objects that result in being adversarial to a DNN. In addition, they showed that more than 99\% of the adversarial poses discovered from InceptionV3 are transferable to other networks such as AlexNet and ResNet-50 trained on ImageNet. 
\\
\indent Beyond Pixel Norm-Balls: Liu et al.~\cite{liu2019pixel} construct adversarial examples through perturbing physical parameters of the image formation model (e.g, Adv. Lighting and Adv. Geometry). The generated images could be used for adversarial data augmentation and creating more `realistic' attacks in the real-world.

\subsection{Differentiable Neural Rendering}
\vspace{-2mm}
Computer graphics based rendering has been known to be non-differentiable, limiting the use of neural networks and back-propagation~\cite{Loper:ECCV:2014}. Recently, there has been some promising work in this area by approximating the rendering pipeline~\cite{kato2020differentiable,kumar2019consistent,kato2017neural,li2019differentiable,tewari2020state}. 
According to Tewari et al.~\cite{tewari2020state},
Neural Rendering (i.e differentiable rendering) is defined as ``Deep image or video generation approaches that enable explicit or implicit control of scene properties such as illumination, camera parameters, pose, geometry, appearance, and semantic structure". Currently, there are four main libraries that allow differentiable rendering: Pytorch3D~\cite{ravi2020pytorch3d}, Kaotlin ~\cite{jatavallabhula2019kaolin}, Tensorflow Graphics, and Mitsuba2~\cite{nimier2019mitsuba}.

Our implementation uses the libraries Pytorch~\cite{NEURIPS2019_9015} and Pytorch3D, which allows using a variety of differentiable rendering algorithms such as Soft Rasterizer~\cite{liu2019softras}. In addition, one of the core design choices of the PyTorch3D API is to support batched outputs. We use batch diffentiable rendering which allows us to first render different views of a single 3D object in a forward pass. Then, allows optimizing from the rendered views (i.e images rendered using different cameras) and back-propagate all the way to the parameters of the mesh. Our implementation will be made available on GitHub.

\section{Assistive Signals Generation}
Most adversarial attacks focus on how to digitally alter natural images~\cite{kurakin2017adversarial}. We apply this approach to modify 2D images in an `assistive' manner. When applied in the 2D space, assistive signals can be used for the creation of \textit{Easier-to-classify} or \textit{Hardened Datasets}. However, our primary interest in this paper is to optimize objects in the 3D space instead of altering images, which is a closer approximation to the physical world. 
In this section, we present how Assistive Signals are generated in different scenarios in both the 2D and 3D spaces. 

\subsection{Assistive Signals in the 2D space}
There is an emerging area related to `robust', `easier-to-classify' and `harder-to-perturb' datasets. As a concrete example, recently Pestana et al.~\cite{Pestana_2021_WACV} proposed a dataset with robust natural images and showed that datasets with robust images are also harder-to-perturb and they could be used to create better adversarial attacks and perform unbiased benchmarking for defenses in the area of adversarial learning. We believe that our work nicely complements~\cite{Pestana_2021_WACV}. They proposed algorithms to find natural robust images, which is an exhaustive process. However, the proposed Assistive Signals allow automatic creation of `easier-to-classify' datasets, which can contribute in this emerging area. 

\begin{algorithm}[h]
  \DontPrintSemicolon
  \caption{Assistive Signals in 2D}
  \KwInput{An input transformation function $\mathcal{F}_{t}$, step size $\alpha$, a classifier $\mathcal{C}$, dataset $\mathcal{D}$, ground truth label $y$, number of iterations $\gamma$}
  \KwOutput{Hardened Dataset $\mathcal{D}_{asst}$}
  \SetKwProg{GenerateHardenedDataset}{GenerateHardenedDataset}{}{}
  \GenerateHardenedDataset{($\mathcal{F}_{t},\alpha,\mathcal{C},\mathcal{D}, y$):}{
  $\mathcal{D}_{asst} \gets \{\}$ \\
  \textbf{for} $x_{n}$ in $\mathcal{D}$ \textbf{do} \\
   \quad$\mathcal{D}_{asst_{n}} \gets \text{Clip}(\mathcal{F}_{t_{a}}(x_{n},\mathcal{C},y,\gamma), 0, 1)$ \\
  \KwRet{$\mathcal{D}_{asst}$}
  }
  \label{alg:dataset-hardening}
\end{algorithm}

\noindent In Algorithm \ref{alg:dataset-hardening}, we define $\mathcal{F}_{t}$ as any iterative targeted objective function (e.g, PGD) such that given a classifier $\mathcal{C}$, an image $x$, ground truth labels $y$ and step size $\alpha$, it generates a perturbed image maximizing the expectation (i.e Binary Cross Entropy Loss) for the ground truth label. Therefore, increasing the confidence score with respect to $y$, which is the opposite aim of adversarial attacks.
\begin{table}[h!]
  \scriptsize
  \centering
  \setlength\tabcolsep{0.75pt}
  \caption{
  First value in every column shows the accuracy (\%) for hardened images using `Assistive Signals' whereas the second value corresponds to the original image. The values are calculated for an ImageNet subset of 5,000 randomly selected images and are tested under two adversarial attacks, FGSM and PGD with different levels or perturbation $\epsilon$. PGD was calculated for 40 iterations.
  }
  \begin{tabular}{ll|l|l|l|l|l|l|}
\cline{3-8}
                                  &                            & \multicolumn{3}{c|}{FGSM}                                                         & \multicolumn{3}{c|}{PGD}                                                          \\ \cline{3-8} 
                                  &                            & \multicolumn{3}{c|}{$\epsilon$}                                                            & \multicolumn{3}{c|}{$\epsilon$}                                                            \\ \hline
\multicolumn{1}{|l|}{Models}      & \multicolumn{1}{c|}{Clean} & \multicolumn{1}{c|}{0.01} & \multicolumn{1}{c|}{0.02} & \multicolumn{1}{c|}{0.04} & \multicolumn{1}{c|}{0.01} & \multicolumn{1}{c|}{0.02} & \multicolumn{1}{c|}{0.04} \\ \hline
\multicolumn{1}{|l|}{Vgg16\_bn}   & 89.8/68.7                  & 71.1/9.3                  & 61.7/6.7                  & 62.8/0.1                  & 59.2/1.4                  & 58.3/1                    & 44.4/0.5                  \\ \hline
\multicolumn{1}{|l|}{ResNet50}    & 98.1/81.2                  & 79.5/14.4                 & 77.2/9.4                  & 64.5/7.4                  & 79.2/8.1                  & 78.1/6                    & 62.7/4.1                  \\ \hline
\multicolumn{1}{|l|}{DenseNet121} & 99.9/87.5                  & 99.9/16.7                 & 82.9/9.3                  & 78.2/6.1                  & 83.9/11.2                 & 75.6/4.5                  & 75.3/3.4                  \\ \hline
\multicolumn{1}{|l|}{InceptionV3} & 95.2/85.9                  & 95.2/13.8                 & 82.4/13.4                 & 57.2/9.2                  & 74.7/12.9                 & 69.8/9.1                  & 57.2/7.2                  \\ \hline
\multicolumn{1}{|l|}{MobileNetV2} & 81.2/72.5                  & 65.5/4.5                  & 43.9/3.8                  & 34.1/3.8                  & 71.7/6.2                  & 72.4/3.1                  & 72.5/2.8                  \\ \hline
\multicolumn{1}{|l|}{SuffleNetV2} & 77.6/80.3                  & 72.3/9.2                  & 55.4/7.1                  & 39.9/6.9                  & 65.2/4.4                  & 70.2/2.9                  & 39.9/1.2                  \\ \hline
\end{tabular}
  \label{tab:assistive_vs_adversarial}%
  \vspace{3mm}
\end{table}%

In Table~\ref{tab:assistive_vs_adversarial}, we show the results of implementing algorithm \ref{alg:dataset-hardening} for ImageNet classifiers. As expected, the accuracy on clean images increases drastically when using assistive signals. However, as a side-effect of making these images more `recognizable', they also get more resilient against adversarial attacks such as FGSM and PGD. Whereas this phenomenon was first studied by Pestana et al.~\cite{Pestana_2021_WACV} for the ImageNet dataset, our proposed method offers a simple but efficient way of creating `Robust' datasets and can be applied to any other datasets.

\begin{figure*}[t!]
    \scriptsize
	\begin{minipage}{0.52\linewidth}
		\includegraphics[width=\linewidth]{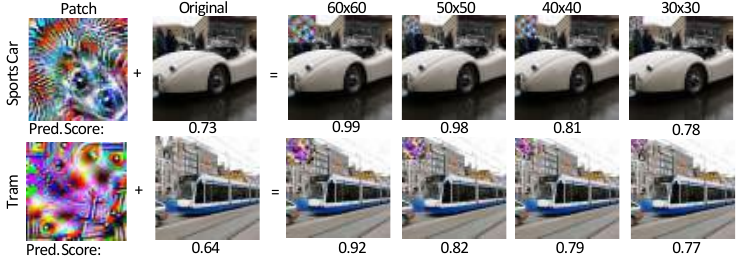}
	\end{minipage}\hfill
	\begin{minipage}{0.47\linewidth}
		\caption{
		 Illustration of \textit{assistive patches} created for the ImageNet classes \textit{Tram}, and \textit{Sports Car} and tested on InceptionV4~\cite{szegedy2016inceptionv4}. The patches were calculated on the target class using random occlusion data augmentations. Once the patches were created, they were resized from 60x60 pixels to 30x30 pixels. The patches are optimized to be location invariant, however, we place them in the top-left corner of the image for visualization. The patch size is written on the top and the prediction confidence score is written below each image.
        }
        \label{fig:patches}
	\end{minipage}
	\squeezeup
\end{figure*}

\subsection{Assistive Signals in the 3D space}
A rendering function $\mathcal{R}$ takes shape parameters $\Theta_{s}$, camera parameters $\Theta_{c}$, lighting parameters $\Theta_{l}$ and texture parameters $\Theta_{t}$ as inputs and outputs a batch of RGB images $\mathcal{I}$. The algorithm uses the batch differentiable rendered images resulting from $\mathcal{R}(\Theta)$ to back-propagate from the ensemble of classifiers all the way back to $\Theta$, which includes the texture $\Theta_{t}$.

Let's define the inputs for $\mathcal{R}$ as $\Theta = \{\Theta_{s},\Theta_{c},\Theta_{l},\Theta_{t}\}$. The function $\mathcal{R}$ computes the gradients of the output image with respect to the input parameters $\frac{\partial \mathcal{I}}{\partial \Theta}$.
While inferring $\Theta_{s}$ and $\Theta_{c}$ from $\mathcal{I}$ is a common task using differentiable renders ~\cite{kato2020differentiable}. To calculate our assistive textures, we will focus on inferring the parameter $\Theta_{t}$. We assume the parameter $\Theta_{t}$ to be either a UV Texture or a Vertex Texture. In addition, $\mathcal{C}$ is a visual classifier. Let $y$ be the true label and $y_{t}$ the target. We use cross-entropy function $\mathcal{L}_{CE}$ as the loss function  for image classification networks such as those trained on ImageNet.
The cross-entropy loss is usually defined as $-{(y\log(p) + (1 - y)\log(1 - p))}$ when number of classes $\mathcal{N}$ is 2. However, for multi-class classification where $\mathcal{N}>2$, the common practice is to calculate a separate loss for each class label per observation and sum the result: $\mathcal{L}_{CE}$=$-\sum_{c=1}^\mathcal{N}y_{c}\log(p_{c})$, where $y_{c}$ is the truth label for the class and $p_{c}$ is the Softmax probability for the $c^{th}$ class. Moreover, a `Clip' function is used to constrain the values in the range [0,1].

\squeezeup
\begin{algorithm}[h!]
  \DontPrintSemicolon
  \caption{Full Texture}
  \KwInput{A render function $\mathcal{R}$, ensemble of classifiers $\mathcal{C}$, 3D mesh/scene parameters $\Theta$, target label $y_{t}$, step size $\alpha$, number of iterations $\gamma$} 
  \KwOutput{texture $\Theta_{t}$}
  \SetKwProg{GenerateTexture}{GenerateTexture}{}{}
  \GenerateTexture{$\mathcal{R}, \mathcal{C}, \Theta, y_{t},\alpha, \gamma$}{
  $\xi \gets 0$\\
  
  \textbf{while} $\xi \leq \gamma$ \textbf{do} \\
    \quad$\mathcal{I} \gets \mathcal{R}(\Theta)$ \\
    \quad L = $\mathcal{L}_{CE}(\mathcal{C}(\mathcal{I}), y_{t})$ \\
    
    \quad$\Theta_{t} \gets \text{Clip}(\Theta_{t} + \alpha \nabla L, 0, 1)$ \\
    
    \quad$\xi \gets \xi + 1$
        
  \KwRet{$\Theta_{t}$}
  \label{alg:assistive-texture}
  }
\end{algorithm}
\squeezeup
Algorithm~\ref{alg:assistive-texture} illustrates the general steps to generate an assistive texture $\Theta_{t}$ targeting the label $y$ for the rendered views $\mathcal{I}$ calculated for a given 3D mesh object. Therefore, given all the parameters needed, the algorithm first generates images from a 3D mesh object using the rendering function $\mathcal{R}$ and parameters $\Theta$, the number of images generated in this step corresponds to the number of cameras passed in $\Theta_{c}$. 
\squeezeup
\begin{algorithm}[h!]
  \DontPrintSemicolon
  \caption{Masked Textures and 3D Patches}
  \KwInput{A render function $\mathcal{R}$, ensemble of classifiers $\mathcal{C}$, 3D mesh/scene parameters $\Theta$, a mask $\mathcal{M}$, target label $y_{t}$, step size $\alpha$, number of iterations $\gamma$} 
  \KwOutput{UV/Vertex texture $\Theta_{t}$}
  \SetKwProg{MaskedTexture}{MaskedTexture}{}{}
  \MaskedTexture{$\mathcal{R}, \mathcal{C}, \Theta, y_{t},\alpha, \gamma$}{
  $\xi \gets 0$\\
  \textbf{while} $\xi \leq \gamma$ \textbf{do} \\
    \quad$\Theta_{t} \gets \text{applyMask}(\Theta_{t},\mathcal{M})$ \\
    \quad$\mathcal{I} \gets \mathcal{R}(\Theta)$ \\
    \quad L = $\mathcal{L}_{CE}(\mathcal{C}(\mathcal{I}), y_{t})$ \\
    
    \quad$\Theta_{t} \gets \text{Clip}(\Theta_{t} + \alpha \nabla L, 0, 1)$ \\
    \quad$\xi \gets \xi + 1$ \\
    \textbf{if} \text{is\_patch}($\Theta$) \textbf{then} \\
    \quad $\mathcal{P} \gets \text{getPatch}(\Theta_{t}, \mathcal{M})$ \\
    \textbf{else} \\
    \quad $\mathcal{P} \gets \text{None}$ \\
    \textbf{endif} \\

  \KwRet{$\Theta_{t}$,$\mathcal{P}$}
  \label{alg:masked-texture}
  }
\end{algorithm}
\squeezeup
While Algorithm~\ref{alg:assistive-texture} applies the perturbations over the whole texture. We refine this algorithm to allow partial perturbations in the object such as \textit{Masked Textures} and \textit{`3D' Patches} in Algorithm~\ref{alg:masked-texture}. $\Theta_{t}$ represents the texture that can be either a UV texture or Vertex texture.  Masked Textures and Patches work similarly in practice, however, we differentiate the patches in that the perturbation is localized in a ``relatively" small area of the 3D object (e.g a car with a small square-shaped assistive patch on the side) whereas Masked Texture could cover the majority of the `car' in this example, but with the exception of windows, front lights, etc. In the 3D space, we can apply a patch similarly to the approach by Brown et al.~\cite{brown2018adversarial} only when we are dealing with a UV Texture. In that case, an extra argument such as a 2D mask with the location of the patch is needed. On the other hand, for a Vertex Texture, every vertex in a mesh can optionally store a RGB color value. For this case, given that the vertices are represented in a list, only a 1D mask is needed.

\section{Experimentation}
\vspace{-2mm}
In this section we perform experiments to test the effectiveness of Assistive Signals for different tasks such as improving the confidence score in 2D images w.r.t their ground truth label, the 'Recognizability' of different 3D objects when using Assistive Textures and Patches, and explaining design features that might contribute to make an object more recognizable to a deep learning classifier (i.e, \textit{Robust Designs}).

\subsection{Assistive Patches}
Adversarial patches~\cite{brown2018adversarial,braunegg2020apricot} have been tested before in the physical world usually by printing the adversarial patterns and then taking pictures of the altered scene containing the patch. One characteristic that makes adversarial patches powerful is the ability of being universal (i.e. they can be applied to any image or scenery). For this reason, adversarial patches are often trained using different background images but optimized for a specific target class. To calculate assistive patches for images (See Figure.~\ref{fig:patches}), we use a similar approach from Brown et al.~\cite{brown2018adversarial}. However, for assistive patches, the objective is to improve the confidence of the ground truth class only, with no (or minimal) effect on images from other classes so as to avoid the misuse of assistive patches (i.e, to become adversarial rather than assistive), which is the opposite goal of Brown et al. and adversarial examples in general. To achieve this , the assistive patch is optimized using images only of the target class. For example, if the optimized assistive patches are targeting the class \textit{motorbike}, we would use only images corresponding to motorbikes. This approach may bring other problems such as overfitting the target class in the training dataset. 

\begin{table}[t!]
  \centering
  \setlength\tabcolsep{1.15pt} 
  \scriptsize
  \caption{Confidence scores (correct class) for five 3D models imaged from three different views where the illumination was also changed for each view (See Fig.~\ref{fig:assistivepatches}).  Using DenseNet121, assistive patches created using conventional methods (optimized with 2D images) \cite{brown2018adversarial} fail or give low confidence for the correct class. However, 3D assistive patches (optimized with 3D neural rendering) always improve the confidence score of the correct class for all views.}
\vspace{1mm}
\begin{tabular}{|l|c|c|c|c|c|c|r|r|r|}
\hline
                  & \multicolumn{3}{c|}{\textbf{Original}}                                                                                                                 & \multicolumn{3}{c|}{\textbf{\begin{tabular}[c]{@{}c@{}}2D Assistive \\ Patches\end{tabular}}}                                                              & \multicolumn{3}{c|}{\textbf{\begin{tabular}[c]{@{}c@{}}3D Assistive\\  Patches\end{tabular}}} \\ \hline
                  & \multicolumn{3}{c|}{Views}                                                                                                                             & \multicolumn{3}{c|}{Views}                                                                                                                             & \multicolumn{3}{c|}{Views}                                                                \\ \hline
\textbf{3D Model} & \multicolumn{1}{l|}{1}                           & \multicolumn{1}{l|}{2}                           & \multicolumn{1}{l|}{3}                           & \multicolumn{1}{l|}{1}                           & \multicolumn{1}{l|}{2}                           & \multicolumn{1}{l|}{3}                           & \multicolumn{1}{l|}{1}       & \multicolumn{1}{l|}{2}       & \multicolumn{1}{l|}{3}      \\ \hline
Airliner          & \multicolumn{1}{r|}{{ 0.59}} & \multicolumn{1}{r|}{{ 0.88}} & \multicolumn{1}{r|}{{ 0.55}} & \multicolumn{1}{r|}{{ 0.57}} & \multicolumn{1}{r|}{{ 0.89}} & \multicolumn{1}{r|}{{ 0.60}}  & { 0.88}  & { 0.95}  & { 0.94} \\ \hline
Traffic Sign      & { x}                         & { x}                         & { x}                         & { x}                         & { x}                         & { x}                         & { 0.83}  & { 0.75}  & { 0.7}  \\ \hline
Car               & \multicolumn{1}{r|}{{ 0.65}} & { x}                         & { x}                         & \multicolumn{1}{r|}{{ 0.73}} & { x}                         & { x}                         & { 0.97}  & { 0.8}   & { 0.71} \\ \hline
Aircraft Carrier  & { x}                         & { x}                         & { x}                         & { x}                         & { x}                         & { x}                         & { 0.67}  & { 0.83}  & { 0.84} \\ \hline
Tram              & \multicolumn{1}{r|}{{ 0.60}}  & \multicolumn{1}{r|}{{ 0.39}} & \multicolumn{1}{r|}{{ 0.47}} & \multicolumn{1}{r|}{{ 0.76}} & \multicolumn{1}{r|}{{ 0.42}} & \multicolumn{1}{r|}{{ 0.46}} & { 0.95}  & { 0.82}  & { 0.54} \\ \hline
\end{tabular}
\label{tab:3dvs2d}
\vspace{2mm}
\end{table}
To avoid overfitting, we use input transformations in the background. In our experiments, one of the most effective and promising transformations tested is the Random Erasing Data Augmentation~\cite{zhong2017random} which applies random occlusion patches to the training data that will make the patch to encode sufficient information about the target class without over-fitting. In addition, for the 2D Assistive Patches (See Table \ref{tab:3dvs2d}) we use the EoT framework (Section 3) to make the patch location invariant.  For this experiment, we collected images from Flickr~\cite{Flickr2020} to prevent data leakage when using the ImageNet for validation. This occurs because of the use of information in the model training process which is not available at prediction time. As a consequence, the predictive scores are overestimated.  Flickr has millions of user shared images. We used the Flickr API\footnote{https://www.flickr.com/services/api/} to download 1,000 images that correspond to objects related to transport vehicles (e.g. bicycles, cars and buses). The API uses a keyword system to download the images. Given that we collected images from the last two years using the Flickr API, and ImageNet was collected in 2012, the probability of repeat images between ImageNet and our collected dataset is extremely low. To filter the new downloaded image labels, we curated the dataset using the pretrained InceptionV3 model. If the prediction from the model is the same as the one used for download (using matching keywords), we retain the image, otherwise, we discard it.
\begin{figure}[t!]
   \centering
   \scriptsize
   \includegraphics[width=\linewidth]{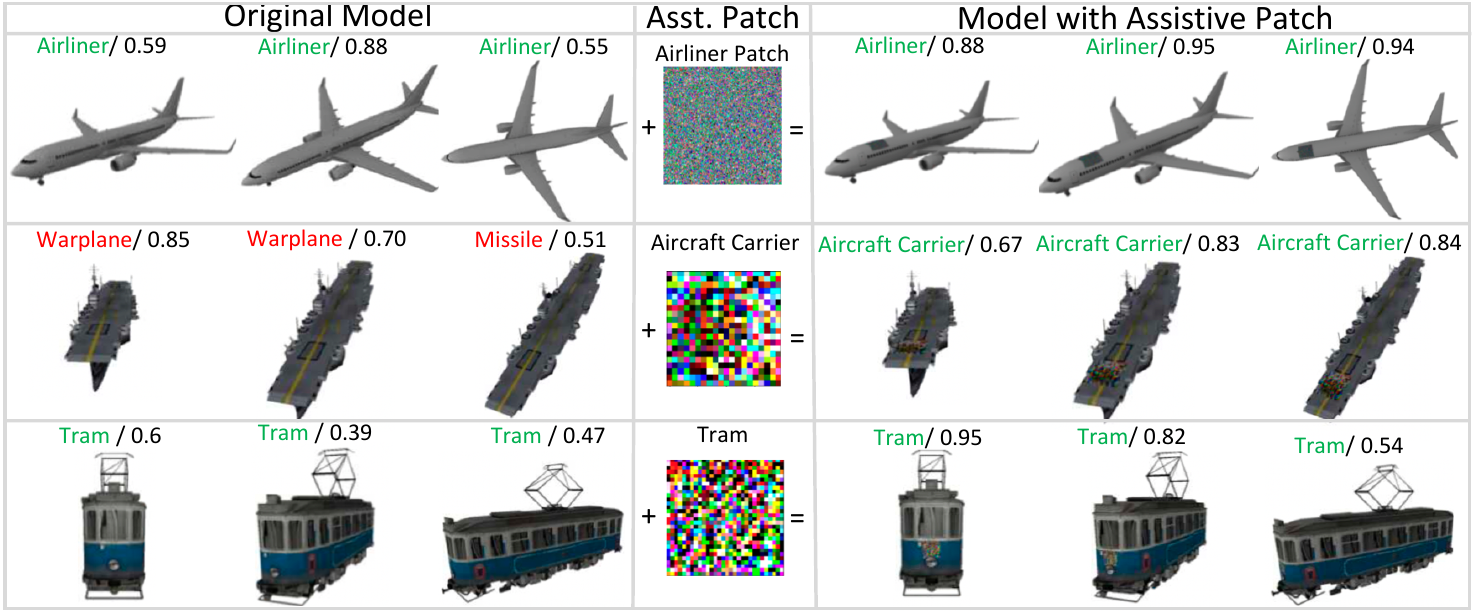} 
   \caption{
   Assistive Patches created for 3 different objects. Each patch was optimized with differentiable rendering using batch rendering sizes of 15, which includes different views of the object and changes in illumination. The patches were optimized for DenseNet121.
   }
\label{fig:assistivepatches}
\vspace{2mm}
\end{figure}

\begin{figure*}[t!]
   \centering
   \scriptsize
   \includegraphics[width=\linewidth]{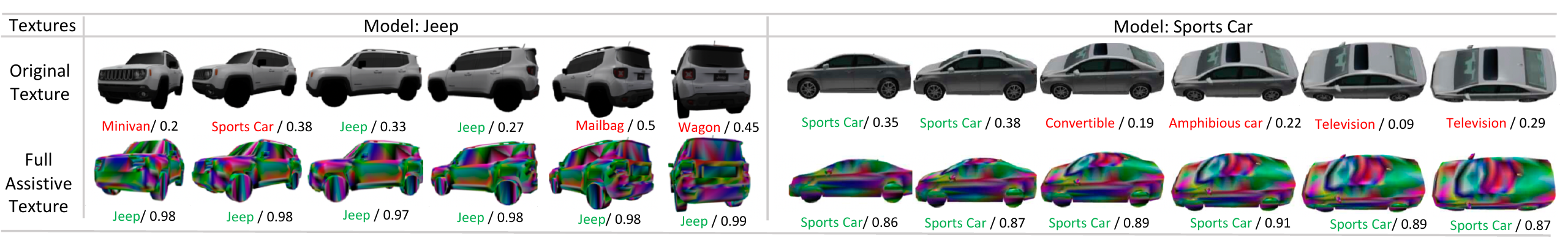} 
   \caption{
   We used algorithm~\ref{alg:assistive-texture} to generate an $\ell_{p}$-unconstrained Full Texture for two different types of car.
   }
   \vspace{-3.5mm}
\label{fig:assistivetextures}
\end{figure*}

In contrast, for the creation of the 3D Assistive Patches we use our approach (See Fig. \ref{fig:differentiable_assistive}), which render views of the 3D object from multiple angles and optimizes the patch end-to-end from the rendered images to the 3d mesh object (i.e, differentiable neural rendering). In Table \ref{tab:3dvs2d} we show that our approach is better than 2D Assistive Patches optimized using the traditional algorithm EoT. Usually, Patches optimized using 2D images only can be location invariant. However, when tested in the real-world by printing the patches, these patches do not transfer well as they do in the digital world. The reason is that conventional 2D Adversarial Patches optimize those without taking into account 3D properties of the scene such as illumination. A more effective approach is simulating the 3D environment and train the patches under different light conditions, cameras, etc. 

In Figure \ref{fig:assistivepatches}, we show different 3D objects and the Assistive Patches created for those objects. Once we apply an Assistive Patch, the confidence score with regards to the 3D object in the scene improves. This holds for different illumination conditions, camera viewpoints or any other 3D scene property that we might be able to optimize using differentiable rendering. 

\subsection{Assistive Textures}
Our method uses Differentiable Neural Rendering, which allows to optimize the texture end-to-end from the rendered images to the 3D mesh object to make it location, lighting, and camera view-point invariant. Figure \ref{fig:assistivetextures} illustrates examples of two 3D objects using Assistive textures where an unconstrained optimization was performed. Given that Assistive Signals are meant to be added to an object on purpose to improve its 'detectability', the perturbations do not need to be `subtle' like in the majority of cases for Deceptive Signals. However, in a more realistic scenario, excluding some parts of the 3D model from the perturbation is desirable (i.e, Masked Textures). Nevertherless, it is important to highlight that unconstrained full body textures reveal semantic meaning encoded in the perturbation, which is what we call \textit{Robust Designs}. We will explain in more detail this concept in the next section.

\begin{figure}[t!]
   \centering
   \scriptsize
   \includegraphics[width=\linewidth]{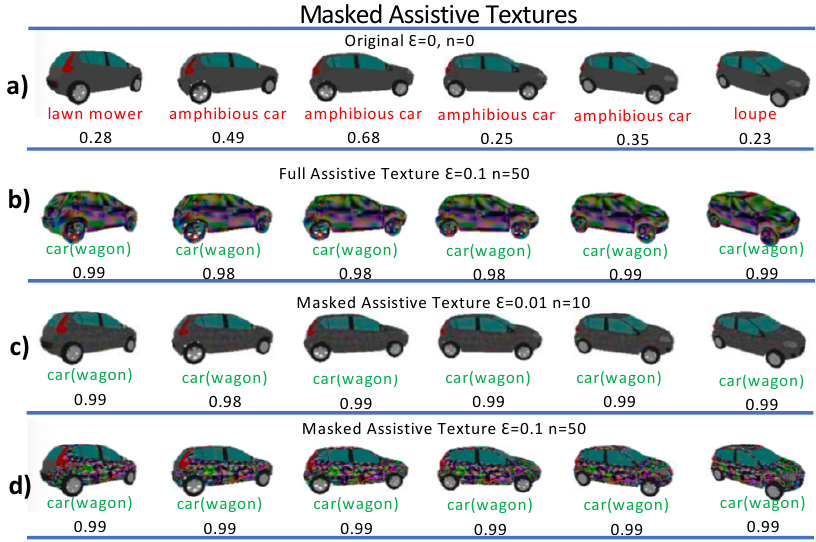} 
   \caption{
   A 3D object from ShapeNetCoreV2~\cite{chang2015shapenet}. We can apply assistive textures (Optimized for all views) to specific regions leaving out the windows, lights and tyres. (a) The original object is misclassified by InceptionV4. (b)  Full assistive texture as shown before in Fig.~\ref{fig:differentiable_assistive}. (c) Masked assistive texture after 10 iterations. (d) Masked assistive texture after 50 iterations.
   }
\label{fig:masked_texture}
\vspace{2mm}
\end{figure}

In the case of masked textures, we can leave out certain regions e.g. the windows, lights and tyres of a car. For instance, Figure \ref{fig:masked_texture} compares an unconstrained full body texture (row b) to \textit{Masked Textures} (row c and d) on a 3D model from the ShapeNetCoreV2~\cite{chang2015shapenet} dataset. The masked assistive textures have been optimized only on specific areas of the car as opposed to its full body perturbation. In the case of row (c), where the perturbation applied is quasi-unnoticeable if compared to the original model (a), the assistive signal improves significantly the right class `car(wagon)'.

Notice how the images in (a) do not contain any details such as the car door handle for example, however, this feature and other details in the door become apparent in (b) (i.e, Robust Design). Another interesting point to note is that full textures tend to be smoother compared to masked textures which appear more `patchy' or noisy. One possible explanation for this phenomenon is that by completely removing the texture from the 3D object (a car), we encourage the full texture optimization to find these `Robust Designs', which are high-level features that are missing in the 3D object to be considered a `car' by the deep-model. In contrast, when creating a masked texture, the texture is partially removed leaving out important features (e.g, windows, tyres) and hence, the algorithm does not need to find those important features anymore and focuses on optimizing other features that look like noisy patterns similarly to conventional crafted perturbations. 

\begin{table}[h!]
  \centering
  \setlength\tabcolsep{1.05pt} 
  \scriptsize
  \caption{
  Transferability of assistive textures generated from different deep models. Correct class confidence scores for a 3D model imaged from three different views where the illumination was also changed for each view are shown. Crosses mean the predicted class was incorrect. First row shows the original object.  $F_{a}$ is a full assistive texture with $\epsilon=0.01$ and $n=50$ iterations.
  $M_{1}$ a is masked assistive textures with perturbation $\epsilon=0.01$ and $n=10$ iterations. $M_{2}$ is masked assistive textures with $\epsilon=0.1$ and $n=50$ iterations.  }
\begin{tabular}{cl|ccc|ccc|ccc|ccc}

                                                                                                                                                                                                \\ \hline
\multicolumn{1}{l}{}                   & \multicolumn{1}{c}{}                                 & \multicolumn{3}{c}{Vgg16}                                             & \multicolumn{3}{c}{ResNet50}                                          & \multicolumn{3}{c}{DenseNet121}                                       & \multicolumn{3}{c}{InceptionV3}                                       \\ \hline
\multicolumn{1}{l}{\multirow{-3}{*}{}} & \multicolumn{1}{c}{\multirow{-3}{*}{Train/Test}} & 1                     & 2                     & 3                     & 1                     & 2                     & 3                     & 1                     & 2                     & 3                     & 1                     & 2                     & 3                     \\ \hline 
                                       
\multirow{-3}{*}{Views}      & Original                                      & x                     & x                     & x                     & x                     & x                     & x                     & x                     & x                     & x                     & { x} & 0.87                  & 0.33                  \\ \hline 
& Vgg16                                                & 0.97                  & 0.98                  & 0.98                  & { x} & 0.39                  & { x} & 0.56                  & 0.43                  & 0.61                  & { x} & 0.14                  & 0.61                  \\
                                       & ResNet50                                             & { x} & 0.1                   & { x} & 0.99                  & 0.98                  & 0.99                  & 0.4                   & 0.74                  & 0.7                   & { x} & 0.6                   & { x} \\
                                       & DenseNet121                                          & { x} & { x} & { x} & 0.13                  & 0.37                  & 0.6                   & 0.99                  & 0.99                  & 0.99                  & { x} & 0.84                  & { x} \\
\multirow{-4}{*}{$F_{a}$}          & InceptionV3                                          & { x} & { x} & { x} & { x} & 0.29                  & 0.09                  & { x} & { x} & { x} & 0.99                  & 0.99                  & 0.99 \\ \hline
                                       & Vgg16                                                & 0.99                  & 0.99                  & 0.99                  & { x} & { x} & { x} & { x} & 0.4                   & 0.79                  & 0.87                  & 0.97                  & 0.17                  \\
                                       & ResNet50                                             & { x} & { x} & { x} & 0.99                  & 0.99                  & 0.99                  & 0.33                  & { x} & 0.21                  & 0.6                   & 0.99                  & 0.69                  \\
                                       & DenseNet121                                          & { x} & { x} & { x} & { x} & { x} & { x} & 0.99                  & 0.99                  & 0.99                  & 0.96                  & 0.99                  & 0.84                  \\
\multirow{-4}{*}{$M_{1}$}          & InceptionV3                                          & { x} & { x} & { x} & { x} & { x} & { x} & { x} & { x} & { x} & 0.99                  & 0.99                  & 0.99                  \\ \hline 
                                       & Vgg16                                                & 0.99                  & 0.99                  & 0.99                  & x                     & 0.08                  & x                     & 0.21                  & { x} & { x} & { x} & { x} & { x} \\
                                       & ResNet50                                             & { x} & { x} & { x} & 0.99                  & 0.99                  & 0.99                  & { x} & { x} & { x} & { x} & { x} & { x} \\
                                       & DenseNet121                                          & { x} & { x} & { x} & { x} & { x} & { x} & 0.99                  & 0.99                  & 0.99                  & { x} & { x} & { x} \\
\multirow{-4}{*}{$M_{2}$}          & InceptionV3                                          & { x} & { x} & { x} & { x} & { x} & { x} & 0.18                  & { x} & { x} & 0.99                  & 0.99                  & 0.99                  
                                         \\ \hline               \label{tab:transfer}%

\end{tabular}
\end{table}

\vspace{-1mm}
\subsection{Transferrability}
\vspace{-2mm}
Deceptive (adversarial) perturbations learned in the 2D image space for one deep-model are known to transfer well to other deep-models~\cite{liu2017delving,huang2020enhancing,demontis2019adversarial}. We tested if the assistive textures optimized using differentible rendering for one deep model are transferable to other deep models. The 3D model itself is kept the same (same car model from Fig.~\ref{fig:masked_texture}). Results are shown in Table \ref{tab:transfer}. \\
We can observe that the full assistive texture $F_{a}$ transfers reasonably well across different CNN-models. We can observe that masked perturbations $M_{1}$ also transfer well across different deep-models and do not need a high magnitude of perturbation $\epsilon$ to achieve this. In contrast, masked perturbations $M_{2}$ have a higher magnitude than $M_{1}$ but transfer poorly to other deep-models.
\vspace{-1mm}
\section{Robust Designs}
\vspace{-3mm}
\begin{figure}[h!]
   \centering
   \scriptsize
   \includegraphics[width=\linewidth]{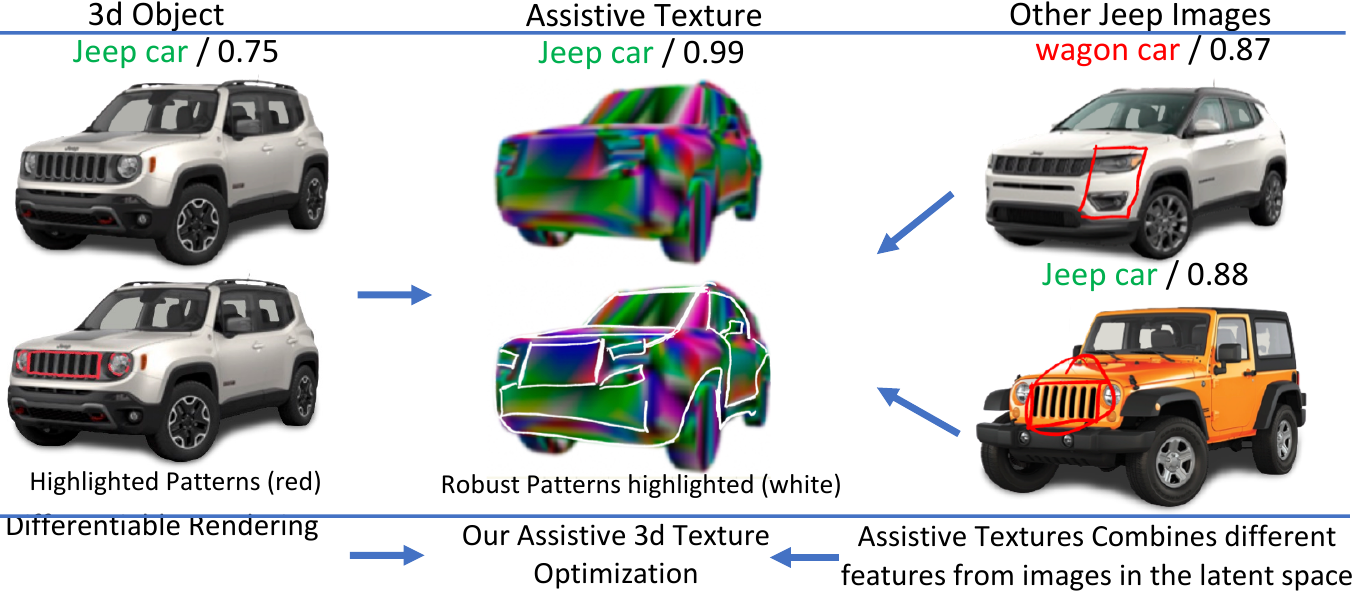} 
   \caption{
   Robust Design example
   }
\label{fig:robustdesign}
\squeezeup
\end{figure}

We found that the patterns created in $\ell_{p}$-unconstrained assistive textures can reveal visual salient patterns, which are human-readable and can be used to provide some insights into what patterns are more detectable to a model. This is what we refer to as \textit{Robust Design}. Figure \ref{fig:robustdesign} illustrates an example of \textit{robust design} that can be used as a method for explainability, but more importantly, it can also provide information about what patterns from the latent space of the trained model are more recognizable to the deep neural network. Given that full textures are highly transferable to other networks, it indicates that this salient patterns are universal an are recognize well by other DNNs.

In some other experiments, we found that they can also provide insights into what colors are more detectable for an specific object. For instance, depending on the country the body of a traffic light can be sometimes gray (metallic), yellow or black. However, the robust designs obtained from different ImageNet models indicate that `black' is the preferable color. This also gives intuitions into what sort of data was used for training an specific algorithm.

In our example, the first column shows an example of a 3D model (jeep car) where we highlighted some of its features. For example, the original rendered image has rounded shaped front lights. Also, we highlighted the grilles from the original model. In the second column, we can observe a rendered image of the same car with a full body assistive texture. After applying our method to extract a full assistive texture from the original 3D object, we can observe that the patterns such as front lights and grilles changed significantly (e.g., having rectangle-shaped front lights instead of rounded). In fact, for this example we found other images corresponding to different models of existing jeep cars outside the ImageNet dataset and we noticed the similarities of the patterns highlighted by the \textit{robust designs} with other jeep models. Observe how in the third column there are two different images of jeep cars. In the process of the texture optimization, the algorithm might find patterns from the latent space that if combined and applied to the texture it improves the confidence score of the targeted class. In the third column we highlighted the features from different jeep cars that are similar to the ones displayed in the assistive texture. Despite the original 3D model having rounded lights and very narrow grilles, in the optimization the assistive signal seems to combine different features from other models.
\vspace{-3mm}

\section{Conclusion}
We introduced the concept of Assistive Signals which, as opposed to Adversarial Examples, aim to improve the confidence score of the ground truth class. At the same time, we provided the first comprenhensive study about the efficacy of Assistive Signals in different scenarios such as 2D images and 3D Textures/Patches. We optimized assistive patches in the 3D space simulating different conditions in the environment such as lighting, cameras, etc. that improved their performance if compared to traditional patches optimized from 2D images. In addition, other than making images or objects more `recognizable' in general, our work explores future directions of assistive signals such as automatic creation of `easier-to-classify' and `Robust' Datasets and explainability through \textit{Robust Designs}. Future directions could extends outside the imaging boundaries and tackle similar problems in other type of data such as Point Clouds, Videos, Sound, etc.
To summarize, while 2D adversarial attacks has been well-studied and technologies such as differentiable rendering allow us to optimize perturbations in 3D models from their projections ~\cite{alcorn2019strike,duan2020adversarial,liu2019pixel,zhang2018camou} (although studied to a lesser extend if compare to 2D images), our approach re-purposes image perturbation for tasks that goes beyond deceptive signals with the aim of fooling a model.

{\small
\bibliographystyle{ieee_fullname}
\bibliography{reference_assistive}
}

\end{document}


\title{Supplementary Material: Assistive Signals for Deep Neural Networks }

\author{Camilo Pestana\\
The University of Western Australia\\
Institution1 address\\
{\tt\small camilo.pestanacardeno@research.uwa.edu.au}
\and
Second Author\\
Institution2\\
First line of institution2 address\\
{\tt\small secondauthor@i2.org}
}

\maketitle
\ificcvfinal\thispagestyle{empty}\fi

\begin{abstract}
   In this supplementary material, we show the algorithms we used to create our assistive signals. In addition, we provide more examples on other applications for assistive signals as well as the potential of our approach to be used for creating deceptive textures.
\end{abstract}
\squeezeup
\squeezeup

\section{`Assistive Signals' terminology}

The term of ‘Assistive Signals’ is not defined in the existing literature as it is a novel research direction. Hence, we coined this term referring to any additive signal that enhances the confidence score of the prediction made by a classifier or recover with ‘good’ confidence the correct
class label under natural or intentional adversarial changes
in the environment. Despite of the similarity that this concept might have with common adversarial defenses, there is a key difference with this two concepts. 
Assistive Signals are meant to improve the general ability of classifiers to predict with the correct class. As explained in our paper, simple changes in the 3d scene such as the pose of the object, camera-view and lightning is enough to fool some models. Assistive Signals tackle this problem by embedding the signals into the 3d object to make it more resilient to this changes. In our threat scenario, we see our 3d simulation as a proxy for real-life scenarios. For that reason, we consider that manipulating the lightning or pose of the object in the real life would be difficult. However, natural adversarial conditions in the scene could cause 

For example, if a 3d object is attacked
by an adversary modifying scene properties such as object
pose or lighting so that it is misclassified by a model, then
‘Assistive Signal’ embedded in the 3d object should ideally
‘assist’ the model to correctly classify the object despite the
adversarial conditions. ‘Assisting’ a ML classifier is the
main purpose of the \textit{Assistive signals}, and such signals can be
expressed in many forms including but not limited to \textit{Assistive
Textures} and \textit{Assistive Patches}. Hence, we believe the term Assistive Signals confers enough meaning to give an intuition about
its purpose and is also generic enough to include all future
works in regards to the many types of assistive signals that
could be created in this emerging area of research.

\section{Easy-to-classify and Hard-to-pertub Images with Assistive Signals}

Dataset hardening is one of the many possible use cases for `Assistive Signals'. As mentioned in our paper, there is an emerging research area related to ‘Robust’, ‘easier-to-classify’ and ‘harder-to-perturb’ datasets. Recently Pestana et al.~\cite{Pestana_2021_WACV} proposed a dataset with robust natural images and showed that datasets that are easier-to-classify accross different ImageNet-based models (called robust images in their paper), are also harder-to-perturb and they could be used
to create better adversarial attacks and perform unbiased
benchmarking for defenses in the area of adversarial learning. Our work complements and extend the work of \cite{Pestana_2021_WACV}. They
proposed algorithms to find natural robust images, which
is an exhaustive process.  Using those exhaustive methods, they were able to create a dataset of robust images for ImageNet with 15k+ images. However, the proposed Assistive
Signals allow automatic creation of robust datasets for any dataset, which
can contribute in this emerging area. 

We performed experiments using different CNN models pretrained on CIFAR10, CIFAR100 and ImageNet. The models include Vgg16~\cite{simonyan2014very}, ResNet50~\cite{he2016deep}, DenseNet121~\cite{huang2017densely}, GoogleNet~\cite{szegedy2014going}, InceptionV3~\cite{szegedy2016rethinking}, ShuffleNetV2~\cite{zhang2017shufflenet,ma2018shufflenet}, MobileNetV2~\cite{howard2017mobilenets,sandler2019mobilenetv2}, and SqueezeNet~\cite{iandola2016squeezenet}. In addition, we used the AdverTorch~\cite{ding2019advertorch} and PyTorch~\cite{paszke2017automatic} libraries for the implementation and creation of assistive signals as well as adversarial attacks used to test the robustness of images ``hardened" with assistive signals. In Table \ref{tab:accuracy-image-specific}, we show the feasibility of using Assistive Signals for the creation of Robust datasets as described by Pestana et al., that are easier to classify as well as harder-to-perturb.

To show the generalization ability of the proposed Assistive signals, we also test them on a non-CNN architecture. We chose transformers for this task as recently, there has been a lot of interest on the potential of using Transformers \cite{vaswani2017attention} (a state-of-the-art technique for natural language processing) for visual classification tasks~\cite{ramachandran2019standalone, wang2020axialdeeplab,carion2020endtoend}. Dosovitskiy et al. \cite{dosovitskiy2020image} introduced Visual Transformers (ViT), which use a conventional Transformer architecture (without Convolutions) and is still able to achieve state-of-the-art performance on different image classification datasets such as ImageNet and ImageNet21k. In Table \ref{tab:accuracy-image-specific}, the results show that transformer-based visual classifiers are also vulnerable to deceptive signals. However, our assistive signals applied are able to improve the predictions of CNN and non-CNN models equally.

\begin{table}[t!]
  \centering
  \setlength\tabcolsep{0.75pt} 
  \scriptsize
  \caption{Accuracy (\%) of deep models on original CIFAR10, CIFAR100 and ImageNet datasets and after adding assistive signals to improve the prediction confidence of the models which can be seen by comparing the accuracy in the two columns labelled ``Clean". With assistive signals, the dataset also becomes more resilient to some weak attacks with small perturbation ($\epsilon$) levels.}
  
    \begin{tabular}{clrrrrrrrr}
    \toprule
    \multirow{3}[6]{*}{DATASET} & \multicolumn{1}{c}{\multirow{3}[6]{*}{MODELS}} & \multicolumn{4}{c|}{Assistive Signals applied} & \multicolumn{4}{c}{Original Dataset} \\
\cmidrule{3-10}          &       & \multicolumn{1}{c}{\multirow{2}[4]{*}{Clean}} & \multicolumn{3}{c}{FGSM} & \multicolumn{1}{c}{\multirow{2}[4]{*}{Clean}} & \multicolumn{3}{c}{FGSM} \\
\cmidrule{4-6}\cmidrule{8-10}          &       &       & \multicolumn{1}{l|}{$\epsilon$=0.01} & \multicolumn{1}{l|}{$\epsilon$=0.02} & \multicolumn{1}{l|}{$\epsilon$=0.04} &       & \multicolumn{1}{l|}{$\epsilon$=0.01} & \multicolumn{1}{l|}{$\epsilon$=0.02} & \multicolumn{1}{l|}{$\epsilon$=0.04} \\
    \midrule
    \midrule
    \multirow{6}[12]{*}{CIFAR10} & Vgg16\_bn & 99.9 & 68.5 & 53.9 & 46.1 & 93.9 & 55.9 & 50.8 & 44.4 \\
\cmidrule{2-10}          & ResNet50 & 99.9 & 78.2 & 64.8 & 57.6 & 93.8 & 57.8 & 50.8 & 45 \\
\cmidrule{2-10}          & DenseNet121 & 99.9 & 73.5 & 65.3 & 60.0 & 94.1 & 63.4 & 50 & 38.9 \\
\cmidrule{2-10}          & Inceptionv3 & 98.9 & 51.8 & 49.0 & 45.7 & 93.7 & 61.3 & 56.9 & 48.4 \\
\cmidrule{2-10}          & GoogleNet & 100 & 85.2 & 73.0 & 56.8 & 92.7 & 39.7 & 32.5 & 25.1 \\
\cmidrule{2-10}          & MobileNetv2 & 99.9 & 86.3 & 86.3 & 39.0 & 94.1 & 34.6 & 27.7 & 19.7 \\
    \midrule
    \midrule
    \multirow{6}[12]{*}{CIFAR100} & Vgg16\_bn & 99.9 & 84.1 & 58.4 & 40.4 & 62.7 & 12.3 & 6.2 & 4.2 \\
\cmidrule{2-10}          & ResNet50 & 99.9 & 90.5 & 64.2 & 38.4 & 72.8 & 14.1 & 6.5 & 4.1 \\
\cmidrule{2-10}          & InceptionV3 & 100 & 90.7 & 69.8 & 37.8 & 75.7 & 16.5 & 10.1 & 6.3 \\
\cmidrule{2-10}          & SqueezeNet & 100 & 88.3 & 57.9 & 28.8 & 69.5 & 6.6 & 6.6 & 6.6 \\
\cmidrule{2-10}          & ShuffleNetv2 & 99.9 & 88.1 & 57.1 & 57.1 & 66.7 & 8.8 & 4.8 & 3.4 \\
\cmidrule{2-10}          & MobileNetV2 & 100 & 82.0 & 51.8 & 21.5 & 65.6 & 8.8 & 6.2 & 5.1 \\
    \midrule
    \midrule
    \multirow{7}[14]{*}{ImageNet} & Vgg16\_bn & 99.6 & 54.9 & 47.7 & 41.3 & 71.8 & 4.2 & 4.3 & 4.7 \\
\cmidrule{2-10}          & ResNet50 & 99.3 & 69.4 & 58.4 & 48.7 & 75.8 & 7.4 & 7.6 & 7.6 \\
\cmidrule{2-10}          & DenseNet121 & 99.9 & 71.8 & 60.5 & 51.4 & 74.6 & 4.2 & 4.4 & 5.6 \\
\cmidrule{2-10}          & InceptionV3 & 99.6 & 70.5 & 70.5 & 46.4 & 69.9 & 13.9 & 13.1 & 13.3 \\
\cmidrule{2-10}          & SqueezeNet & 99.9 & 62.1 & 49.7 & 34.6 & 58.1 & 0.7 & 0.6 & 0.7 \\
\cmidrule{2-10}          & ShuffleNetv2 & 100 & 58.2 & 49.3 & 35.5 & 69.4 & 2.7 & 3.2 & 3.8 \\
\cmidrule{2-10}          & MobileNetV2 & 99.7 & 55.6 & 48.4 & 38.3 & 72.5 & 1.3 & 2.0 & 3.5 \\
\cmidrule{2-10}          & \textbf{ViT\_B\_16} & 100 & 65.9 & 55.0 & 46.7 & 79.5 & 13.8 & 10.8 & 10.8 \\
\cmidrule{2-10}          & \textbf{ViT\_B\_32} & 100 & 67.6 & 53.8  & 44.2 & 75.6 & 20.4 & 14.4 & 12.3  \\
\cmidrule{2-10}          &
\textbf{ViT\_L\_16}  & 100 & 71.5 & 58.2   & 48.0 & 81.1 & 23.2 & 20.3  & 19.7  \\
\cmidrule{2-10}          &
\textbf{ViT\_L\_32} & 100 & 73.8 & 61.4  & 51.5 & 76.0 & 23.6 & 19.7 & 19.8  \\

    \bottomrule
    \end{tabular}%
  \label{tab:accuracy-image-specific}%
  \squeezeup
  \squeezeup
\end{table}%

\section{Algorithms}

Algorithm 1 returns a dataset with hardened images. In the process of hardening, we pass as a parameter s strong targeted  adversarial attack (e.g, PGD).

\begin{algorithm}[h]
  \scriptsize
  \DontPrintSemicolon
  \caption{Image Specific Assistive Signal}
  \KwInput{Correct targeted classes $k$, iterative targeted adversarial attack $atk$, a learning rate per step $lr$, a maximum number of iterations $N$, dataset $D$, model $M$} 
  \KwOutput{A hardened dataset $D_{asst}$}
  \SetKwProg{AssistiveSignal}{AssistiveSignal}{}{}
  \AssistiveSignal{($k,atk,lr,D, M$)}{
  \tcp{Initialize New Dataset} 
  $D_{asst} \gets []$ \\
    \tcp{For each image $i$ in dataset $D$}
    \ForEach{$x \in D$}{ 
         \tcp{Generate Assistive Image} 
          $i_{asst} \gets  \text{clip}(\text{atk}(i, k, lr, N), 0, 1)$ \\ 
          $D_{asst}$.insert($i_{assist}$) \\
    }

  \KwRet{$D_{asst}$}
  }
\end{algorithm}

Algorithm 2 shows the process for creating an assistive patch in 2d images.

\begin{algorithm}[h]
      \scriptsize
      \DontPrintSemicolon
      \caption{Assistive Signal Patch}
      \KwInput{A target class $k$, a learning rate $lr$, patch size $p_{size}$, a maximum of iterations $N$, dataset $D$, model $M$} 
      \KwOutput{An Assistive Patch $p$}
      \SetKwProg{AssistivePatch}{AssistivePatch}{}{}
      \AssistivePatch{($cls,lr, p_{size}
      ,D, M$)}{
      \tcp{Initialize Applied Patch} 
      $applied \gets applyPatch(p_{size},l_{x}, l_{y})$ \\
      \ForEach{$n \in N$}{
        \tcp{For each image $i$ in dataset $D$}
        \ForEach{$i \in D$}{
              $l_{x}, l_{y} \gets randomLocation(p_{size})$ \\
              $mask  \gets getMask(p_{size},l_{x}, l_{y})$ \\
              \tcp{Generate Assistive Image}
              $i_{asst} \gets mask\times applied + (1-mask)\times i$ \\
              $applied \gets clip(lr \times \nabla i_{asst} + applied, 0, 1) $
              
              $i_{asst}\gets applyTransforms(applied)$\\
              $p = getPatch(i_{asst})$
        }
       }
      
      \KwRet{$i_{asst}, p$}
      }
\end{algorithm}

Algorithm 3 shows how to create an assistive signal patch optimized from a 3d object. In this process, we use a 3d object using UV textures which is passed along with the size and location where the patch will be applied in the UV texture image. From there, different camera and illumination parameters can be passed for the process of batch neural rendering. We found that optimizing the patch using different cameras views at the same time usually creates a better patches that are able to stand changes different changes in the scene. However, depending on the 3d scene parameters and the object itself, the hyperparameter of number of views can make a significant difference in the training of the patch.

\begin{algorithm}[h]
      \scriptsize
      \DontPrintSemicolon
      \caption{Assistive 3D Patch}
      \KwInput{A target class $k$, pertubation level $\epsilon$, batch size $bs$, location of the patch $loc$, size of patch $sp$, 3d scene parameters $params$, a maximum of iterations $N$, 3d object $obj$, deep model $M$} 
      \KwOutput{An Assistive Patch $p$}
      \SetKwProg{AssistiveDiffRendPatch}{AssistiveDiffRendPatch}{}{}
      \AssistiveDiffRendPatch{($cls,lr, p_{size}
      ,D, M$)}{
      \tcp{Get UV Texture from 3d object}
      $uv$ = $\text{getTexture}(obj)$ \\
      \tcp{Initialize patch}
      $p$   = random($sp$) \\ 
      \ForEach{$n \in N$}{
          \tcp{Apply patch to texture}
          $applied_{p}$, $mask$ = applyPatch($uv$, $patch$, $loc$).  \\
          \tcp{Batch neural rendering} 
          images = render($obj$,$params$, $bs$) \\
          \ForEach{$r_i \in images$}{
            $i_{asst} \gets clip(lr \times \nabla r_i + applied, 0, 1)$
          }
           $p = \text{getPatch}(i_{asst}, mask)$
        
       }
      
      \KwRet{$i_{asst}, p$}
      }
\end{algorithm}

Similarly to Algorithm 3, Algorithm 4 optimized a vertex texture of a 3d object from a batch of rendered images. However, the original RGB values of the vertex textures are replace by the values (0.5,0.5,0.5) in the initialization. Another variation of this algorithm is the \textit{Masked Texture} where we combine algorithm 3 and 4 to create an assistive texture that has masked area in the texture. 
\begin{algorithm}[!t]
      \scriptsize
      \DontPrintSemicolon
      \caption{Assistive 3D Texture}
      \KwInput{A target class $k$, pertubation level $\epsilon$, batch size $bs$, 3d scene parameters $params$, a maximum of iterations $N$, 3d object $obj$, deep model $M$} 
      \KwOutput{An Assistive Patch $p$}
      \SetKwProg{AssistiveDiffRendPatch}{AssistiveDiffRendPatch}{}{}
      \AssistiveDiffRendPatch{($cls,lr, p_{size}
      ,D, M$)}{
      \tcp{Initialize Vertex Texture}
      $v$ = $\text{setTexture}(obj, (0.5,0.5,0.5)$ \\
      \ForEach{$n \in N$}{
         \tcp{Batch neural rendering} 
          images = render($obj$,$params$, $bs$) \\
          \ForEach{$r_i \in images$}{
            $i_{asst} \gets clip(lr \times \nabla v + r_{i}, 0, 1)$
          }
        
       }
      
      \KwRet{$i_{asst}, obj$}
      }
\end{algorithm}

\begin{figure*}[h]
   \centering
   \includegraphics[width=\linewidth]{images/assistive/deceptives.pdf} 
   \caption{
    Deceptives textures optimized for two different 3d objects using our neural rendering method (Algorithm 4).
   }
\label{fig:deceptives}

\end{figure*}

\comment{
\vfill
  \begin{center}    
      \includegraphics [width=5 in]{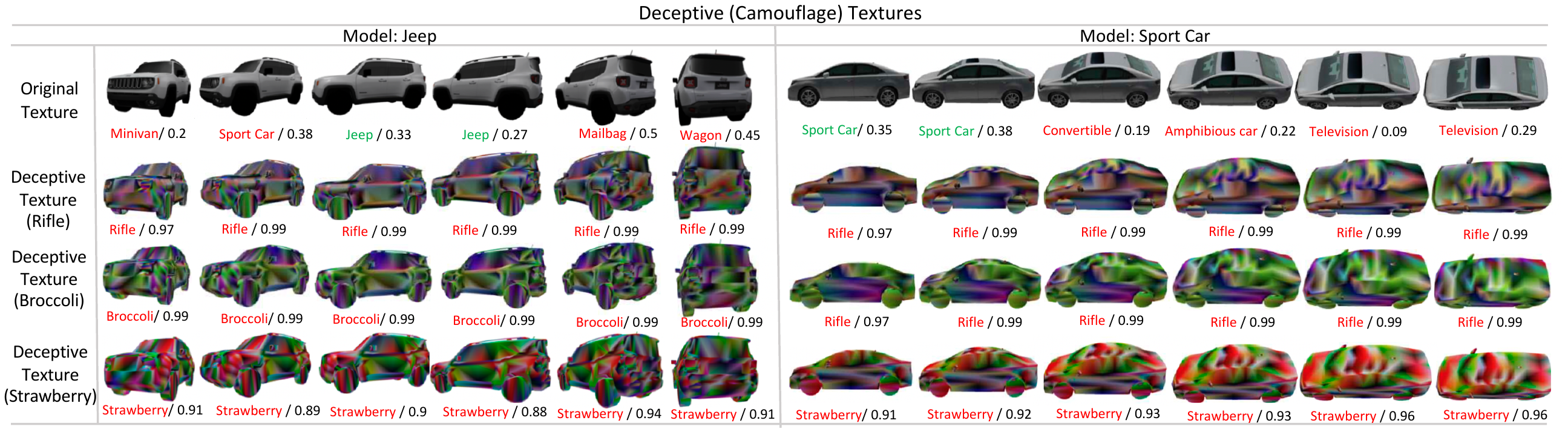}
  \end{center}
\vfill
}

\section{Examples of Deceptive Textures}

Despite of focusing on assistive textures, in Figure \ref{fig:deceptives} we show examples to illustrate the potential of our approach to create deceptive (adversarial) textures in 3d objects that can be invariant to some 3d scene parameters. In this example, we optimized the model from 10 different views (we show only the key images in the interpolation) at the same time. Other than using different camera views, we also trained the textures using different illumination conditions. All 3d objects with our deceptive textures are able to successfully fool a deep model from different camera views and illumination conditions.

{\small
\bibliographystyle{ieee_fullname}
\bibliography{reference_assistive}
}